\documentclass[11pt]{article}

\usepackage[preprint]{acl}

\usepackage{times}
\usepackage{latexsym}

\usepackage[T1]{fontenc}

\usepackage[utf8]{inputenc}

\usepackage{microtype}

\usepackage{inconsolata}

\usepackage{graphicx}

\usepackage{microtype}
\usepackage{graphicx}
\usepackage{subcaption}
\usepackage{booktabs} 

\usepackage{hyperref}

\usepackage{xspace}
\newcommand{\methodname}{StiefAttention\xspace}
\usepackage{amsmath}
\usepackage{amssymb}
\usepackage{mathtools}
\usepackage{amsthm}
\usepackage{multirow}
\usepackage[capitalize,noabbrev]{cleveref}
\usepackage{algorithm}
\usepackage{algorithmic}

\title{Don't be so Stief!\\Learning KV Cache low-rank approximation over the Stiefel manifold}

\author{
 \textbf{Luca Benfenati\textsuperscript{1}},
 \textbf{Matteo Risso\textsuperscript{1}},
 \textbf{Andrea Vannozzi\textsuperscript{1}},
 \textbf{Ahmet Caner Yüzügüler\textsuperscript{2}},
\\
 \textbf{Lukas Cavigelli\textsuperscript{2}},
 \textbf{Enrico Macii\textsuperscript{1}},
 \textbf{Daniele Jahier Pagliari\textsuperscript{1}},
 \textbf{Alessio Burrello\textsuperscript{1}}
\\
\\
 \textsuperscript{1}Department of Control and Computer Engineering, Politecnico di Torino, Italy,\\
 \textsuperscript{2}Huawei Zurich Research Center, Switzerland
\\
 \small{
   \textbf{Correspondence:} \href{mailto:luca.benfenati@polito.it}{luca.benfenati@polito.it}
 }
}

\begin{document}
\maketitle
\begin{abstract}
Key-value (KV) caching enables fast autoregressive decoding but at long contexts becomes a dominant bottleneck in High Bandwidth Memory (HBM) capacity and bandwidth.
A common mitigation is to compress cached keys and values by projecting per-head matrices to a lower rank, storing only the projections in the HBM.
However, existing post-training approaches typically fit these projections using SVD-style proxy objectives, which may poorly reflect end-to-end reconstruction after softmax, value mixing, and subsequent decoder-layer transformations.

For these reasons, we introduce \methodname, a post-training KV-cache compression method that learns \emph{orthonormal} projection bases by directly minimizing \emph{decoder-layer output reconstruction error}.
\methodname additionally constructs layer-wise error-rank profiles over candidate ranks, enabling sequential rank allocation under a user-specified KV cache budget.
Notably, on Llama3-8B under the same conditions, \methodname outperforms EigenAttention by $4.2$ points on C4 perplexity and $8.9$ points on 0-shot MMLU accuracy at iso-compression, yielding lower relative error and higher cosine similarity with respect to the original decoder-layer outputs.
\end{abstract}

\section{Introduction}
\label{sec:intro}
Large language models (LLMs)~\cite{touvron2023llama2openfoundation, grattafiori2024llama3herdmodels} achieve strong performance across a wide range of language tasks, and are increasingly deployed in interactive and long-context settings.
Yet, long-context inference is often bottlenecked by memory capacity and bandwidth.

In particular, during autoregressive decoding, attention layers store the key and value projections of past tokens in a Key Value (KV) cache, avoiding their recomputation for each newly generated token.

While this greatly reduces compute, the KV cache size grows linearly with both sequence length and batch size~\cite{scalingLLM}: for example, even for a compact model such as Llama3-8B with Grouped-Query Attention (GQA), caching keys and values in half-precision at a relatively short context length $32{,}768$ and assuming a batch size of $4$ requires $\approx 16$\,GB, comparable to the model's fp16 weights footprint. This pressure becomes more pronounced as context lengths continue to increase~\cite{openrouter}, potentially exceeding available High Bandwidth Memory (HBM) capacity and bandwidth.

To mitigate the memory-occupancy issues, high data-transfer latency, and energy consumption associated with KV caching, one approach is to reduce KV cache memory \emph{by design} via architectural changes that cache low-dimensional latent states rather than full keys and values.
Multi-Head Latent Attention (MLA) is a representative example~\cite{deepseekv2}, which however requires re-training to adapt to the modified attention mechanism.
TransMLA~\cite{transmla} converts a model to MLA post-training, yet still relies on a fine-tuning stage for adaptation.

In this paper, we focus instead on post-training methods that operate on a fixed pre-trained model.
Within this category, several optimization axes have been explored, including quantization~\cite{hooper2024kvquant, LiuIcml24}, token eviction~\cite{streamingLLM, h2o}, and context sharing~\cite{ZhengNips24}. Orthogonal to these methods, recent work focuses on compressing the embedding dimension of the KV cache~\cite{asvd, chang2025palu, saxena2024eigenattention}. This is achieved by applying low-rank matrix decomposition techniques, such as Singular Value Decomposition (SVD), to reduce KV cache size, with an objective function that minimizes the compression error of key and value vectors. Although these methods reduce KV cache size by up to 1.67$\times$~\cite{saxena2024eigenattention}, they often suffer from significant accuracy drops, which hinder their effectiveness.

In this paper, we argue that the accuracy degradation observed in SVD-based compression techniques primarily stems from their proxy optimization objectives. 
Minimizing reconstruction error for each cached key or value vector can yield low per-vector distortion, but it does not model how these distortions interact with the attention softmax, value mixing, and subsequent decoder-layer computations (normalization, residual pathways, and nonlinearities).
Consequently, a projection that is optimal under a proxy reconstruction objective can still induce a large decoder-layer output error, which compounds across depth and ultimately degrades end-to-end generation quality.

To address this mismatch, we propose \textit{\methodname}, a post-training KV cache compression method that \textbf{minimizes error directly at the decoder layer output}. Our method trains a lightweight predictor using activation statistics to find optimal orthonormal projection bases lying on the Stiefel manifold~\cite{stella}. At inference time, keys and values are stored in compressed form, resulting in a significant reduction in HBM footprint and bandwidth requirements. We evaluate {\methodname} on Llama3-8B and compare it to EigenAttention~\cite{saxena2024eigenattention} under the same KV budget, finding improved end-to-end quality and better layer-output reconstruction quality under KV cache compression.

Our contributions are: i) we study KV cache compression along the per-head feature dimension under a layer-wise \emph{decoder-layer output} reconstruction error budget; ii) we introduce \textit{\methodname}, which learns orthonormal bases from activation statistics and improves end-to-end quality at matched KV cache budgets (e.g., $-8.9$ C4 perplexity and $+4.2$ MMLU points vs.\ EigenAttention); iii) we provide analyses showing that optimizing decoder-layer outputs improves directional consistency ($+3.3\%$ cosine similarity and $5.2\%$ lower layer-output error), even though SVD-style methods better reconstruct attention outputs ($29.6\%$ lower attention-output error).

\section{Background}
\label{sec:background}

\subsection{Compression along the head dimension}
We compress the KV cache by projecting along the per-head feature dimension $d_h$.
For a key matrix $K\in\mathbb{R}^{n\times d_h}$, let
$P_K\in\mathbb{R}^{d_h\times r_K}$ be a column-orthonormal projection basis, with $r_K\ll d_h$.
The compressed key cache is
\begin{equation}
K^{\downarrow}=K P_K\in\mathbb{R}^{n\times r_K},
\label{eq:bg_compress_k}
\end{equation}
and the reconstructed approximation used in attention is
\begin{equation}
\tilde K = K^{\downarrow}P_K^\top = K P_K P_K^\top.
\label{eq:bg_reconstruct_k}
\end{equation}
The same construction is applied to values using
$P_V\in\mathbb{R}^{d_h\times r_V}$, yielding
$V^{\downarrow}=V P_V$ and $\tilde V=V P_V P_V^\top$.
The resulting per-token KV-cache ratio is
\begin{equation}
\mathrm{CR}=\frac{r_K+r_V}{2d_h}.
\label{eq:bg_cr}
\end{equation}
Throughout the paper, $P_K$ and $P_V$ denote generic projection bases lying on the Stiefel manifold, i.e., the set of matrices with orthonormal columns~\cite{stella}.

\subsection{Projection-based baselines and proxy objectives}
\label{subsec:backg_svdkvc}
Projection-based KV-cache compression methods use the scheme in Eqs.~\eqref{eq:bg_compress_k}--\eqref{eq:bg_cr}, but differ in the \emph{proxy objective} used to choose the bases.

\paragraph{Reconstruction-based objectives.}
A standard SVD baseline chooses $P_K$ to minimize the Frobenius reconstruction error of $K$, and applies the same construction to $V$~\citep{asvd,chang2025palu}.
This preserves the cached tensors themselves, but it does not directly optimize attention scores, value mixing, or decoder-layer outputs.

\paragraph{EigenAttention.}
EigenAttention~\citep{saxena2024eigenattention} observes that projecting keys also induces an implicit projection on queries in the attention score computation $QK^\top$.
It therefore forms the vertically concatenated matrix
\begin{equation}
Z=\begin{bmatrix}K\\Q\end{bmatrix}\in\mathbb{R}^{2n\times d_h},
\label{eq:bg_eigen_concat}
\end{equation}
and chooses a shared orthonormal basis $P_K\in\mathbb{R}^{d_h\times r_K}$ by minimizing
\begin{equation}
\min_{P_K^\top P_K=I_{r_K}}\;\;\|Z-ZP_KP_K^\top\|_F^2,
\label{eq:bg_eigen}
\end{equation}
which is solved by the truncated SVD of $Z$.
This remains a reconstruction objective: because the SVD is computed on $[K;Q]$, the learned subspace is influenced by the relative scale of $K$ and $Q$, and the higher-energy block can dominate the top singular directions.
For values, EigenAttention uses the same reconstruction objective as K-SVD.

\paragraph{Interaction-based objectives.}
KQ-SVD~\citep{lesens2025kqsvd} instead targets the pre-softmax interaction matrix $QK^\top$, deriving low-rank factors that approximate attention scores directly.
For values, it mirrors the same idea on the value--output pathway by approximating $V W_O$ in low rank, where $W_O$ is the attention output projection.
Although this better matches attention interactions than tensor reconstruction, it still optimizes an intermediate proxy rather than the full decoder-layer output.
Additional notation and detailed objective forms are reported in Appendix~\ref{app:background_details}.
\section{Method: \methodname}
\label{sec:method}
\begin{figure}
    \centering
    \includegraphics[width=\linewidth]{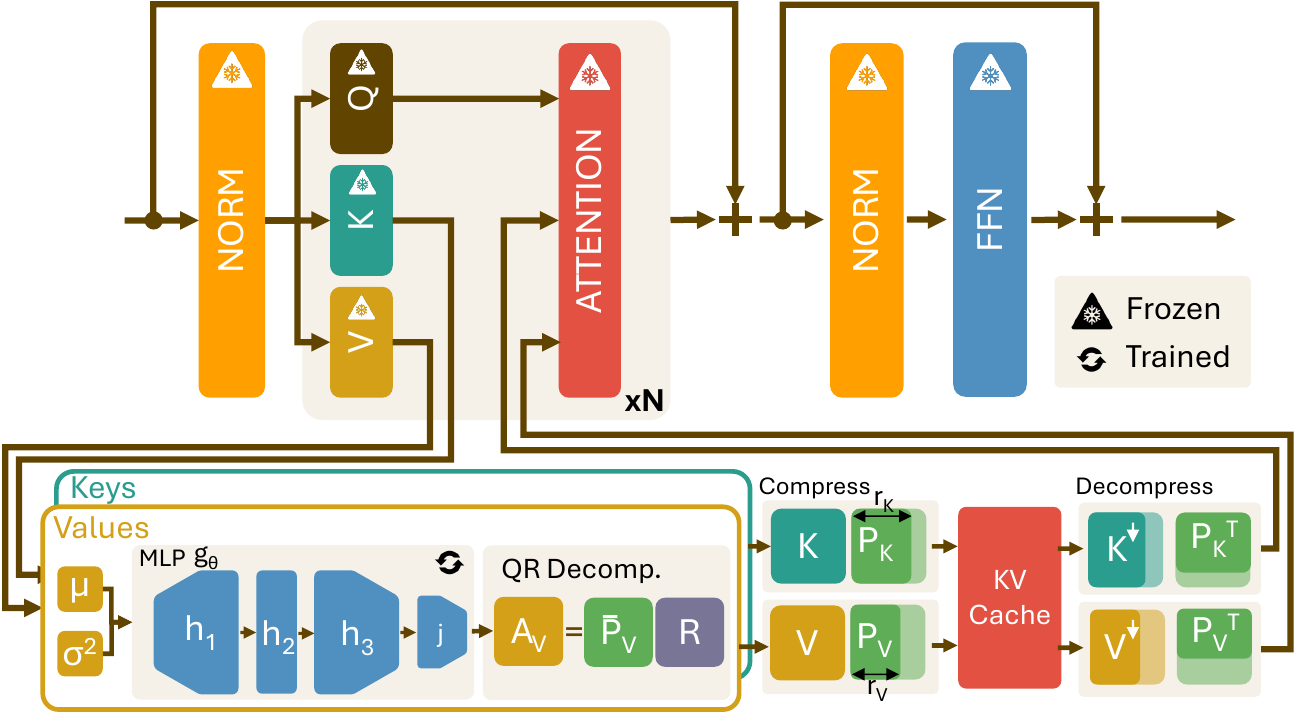}
    \caption{Overview of \methodname. From lightweight activation statistics $\mu_{K}, \mu_{V}$ and $\sigma_{K}^2, \sigma_{V}^2$, we learn orthonormal projection bases $P_K, P_V$ which are optimized to minimize decoder-layer output error.}
    \label{fig:stief_att}
\end{figure}

\subsection{Goal}
SVD-like methods introduced in Sec.~\ref{subsec:backg_svdkvc} optimize proxy objectives on intermediate quantities (e.g., $K$, $[K;Q]$, or $QK^\top$).
In contrast, \methodname learns low-rank KV projections that directly minimize \emph{full decoder layer output} error.
Fig.~\ref{fig:stief_att} summarizes the core idea: instead of optimizing reconstruction in KV space, we optimize the error measured at the decoder-layer output, capturing the effect of both softmax and value mixing, as well as downstream transformations in the decoder layer, which includes output projection, residual paths, normalization and the MLP.

\subsection{Problem statement}
For each decoder layer $\ell$, let $f_\ell$ denote the original layer mapping and let $\tilde{f}_\ell$ denote the layer where keys and values inside attention are replaced by their reconstructed low-rank forms. For readability, we first write the compression for a single KV head and omit the value-head index; the head-sharing scheme used in the implementation is described in Sec.~\ref{par:head_sharing}.
\methodname learns the $\bar P_K\in\mathbb{R}^{d_h\times d_h}$ and $\bar P_V\in\mathbb{R}^{d_h\times d_h}$ \emph{orthonormal projection bases}, and uses their leading $r_K, r_V \ll d_h$ columns to define the corresponding low-rank projection matrices:
\begin{equation}
\begin{aligned}
&P_K^{(r_K)} \coloneq \bar P_K[:,1{:}r_K]\in\mathbb{R}^{d_h\times r_K},\\
&P_V^{(r_V)} \coloneq \bar P_V[:,1{:}r_V]\in\mathbb{R}^{d_h\times r_V},
\end{aligned}
\label{eq:dkv_trunc_def}
\end{equation}
with $(P_K^{(r_K)})^\top P_K^{(r_K)} = I_{r_K}$ and $(P_V^{(r_V)})^\top P_V^{(r_V)} = I_{r_V}$.

Given full-precision matrices $K,V\in\mathbb{R}^{n\times d_h}$ (as obtained during calibration, or before compression is applied), \methodname forms and caches the compressed tensors
$K^{\downarrow}=K P_K^{(r_K)}$ and $V^{\downarrow}=V P_V^{(r_V)}$.
Whenever the attention block needs $K$ and $V$ in the original dimension, it uses the reconstructed forms:
\begin{equation}
\begin{aligned}
&\tilde K = K^{\downarrow}(P_K^{(r_K)})^\top = K P_K^{(r_K)}(P_K^{(r_K)})^\top,\\
&\tilde V = V^{\downarrow}(P_V^{(r_V)})^\top = V P_V^{(r_V)}(P_V^{(r_V)})^\top.
\end{aligned}
\label{eq:dkv_reconstruct}
\end{equation}

Given a set of calibration inputs $x$, \methodname finds $P_K$ and $P_V$ by solving through gradient descent the optimization problem:
\begin{equation}
\min_{\theta}\;\; \Delta_\ell(r_K,r_V;\theta;x),
\label{eq:dkv_obj}
\end{equation}
where $\Delta_\ell(r_K,r_V;\theta;x)$ denotes the relative error at the decoder-layer output:
\begin{equation}
\small
\Delta_\ell(r_K,r_V;\theta;x)
=
\mathbb{E}_{x}\!\left[
\frac{\| f_\ell(x) - \tilde{f}_\ell(x; r_K,r_V,\theta) \|_F}{\| f_\ell(x) \|_F}
\right],
\label{eq:dkv_delta}
\end{equation}
and $\theta$ parameterizes the predictor from which the projection bases $(P_K,P_V)$ are derived, described in the next Sec.~\ref{subsec:methods_basis_prediction}.

\subsection{Gradient-based basis prediction}
\label{subsec:methods_basis_prediction}
\methodname computes the $P_K, P_V$ orthonormal projection bases from simple activation statistics.
Let $K_x$ denote a collection of key vectors for layer $\ell$ computed over calibration samples $x$ and previously cached tokens.
We compute per-dimension mean and variance:
\begin{equation}
\begin{aligned}
&\mu_K = \mathbb{E}[K_x] \in \mathbb{R}^{d_h},\\
&\sigma_K^2 = \mathbb{E}\!\left[(K_x-\mu_K)^2\right] \in \mathbb{R}^{d_h},
\end{aligned}
\label{eq:dkv_stats}
\end{equation}
which are aggregated to form features $s_K=[\mu_K;\sigma_K^2]\in\mathbb{R}^{2d_h}$.
The same computation is performed to compute $s_V$ features over $V$ vectors.
Intuitively, $s_K$ and $s_V$ provide a cheap summary of first/second-order activation statistics from which the projection bases are computed, as will be described in subsequent paragraphs.
\paragraph{MLP architecture.}
We define a trainable predictor $g_\theta$ based on the MLP architecture depicted in the bottom part of Fig.~\ref{fig:stief_att}. The same architecture is employed for both keys and values. The predictor $g_\theta$ is trained by solving through gradient descent the optimization problem of Eq.~\ref{eq:dkv_obj}, to map the $s_K$ and $s_V$ statistics to $A_K$ and $A_V$ matrices whose orthonormalization and rank approximation yield a basis that minimizes decoder-layer output error.

Concretely, $g_\theta$ is an MLP composed of three hidden layers $h_i$ followed by a linear head $j$ that outputs the pre-orthonormalization matrix $A \in \mathbb{R}^{d_h\times d_h}$.
Let $s\in\mathbb{R}^{2d_h}$ denote the input feature vector (i.e., $s=s_K \: \text{or} \: s_V$).
The general form of hidden layers $h_i$ is:
\begin{equation}
h_i = \phi \left( \mathrm{LN}\left(W_i h_{i-1} + b_i\right) \right) \ \text{with} \ i \in \left[1, 3\right]
\label{eq:dkv_mlp}
\end{equation}
where $h_0 \equiv s$, $W_i$ and $b_i$ are the learnable weights and biases generically denoted as $\theta$ in Eq.~\ref{eq:dkv_obj} and Eq.~\ref{eq:dkv_delta}. $\phi(\cdot)$ is the GELU non-linearity and $\mathrm{LN}$ denotes LayerNorm.
While the final linear head producing $A$ is simply $j = W_j h_3 + b_j$.
We finally define:

\begin{equation}
A_K = g_{\theta_K}(s_K),\qquad A_V = g_{\theta_V}(s_V).
\label{eq:dkv_raw}
\end{equation}
where $\theta_K$ and $\theta_V$ respectively define the set of trainable parameters $\theta$ specific for the separate Key and Value predictors.

\paragraph{Orthonormalization via QR.}
We orthonormalize $A_K$ and $A_V$ via QR decomposition~\cite{bookqr}:
\begin{equation}
A_K = Q_K R_K,\qquad A_V = Q_V R_V,
\label{eq:dkv_qr}
\end{equation}
where $Q_K,Q_V\in\mathbb{R}^{d_h\times d_h}$ are matrices with orthonormal columns and $R_K,R_V$ are upper-triangular.\footnote{Here $Q$ does not denote the query vectors of Eq.~\ref{eq:bg_attn}.}
In \methodname, we use only the orthonormal factor $Q$, because our compression/reconstruction depends on the orthogonal projector $P P^\top$, which is determined solely by the subspace spanned by the columns of $P$.
Since $A$ and $Q$ span the same column space in a QR factorization, $Q Q^\top$ projects onto the same subspace as $A$.
The triangular factor $R$ only encodes a change of coordinates (and scaling) within that subspace and is therefore irrelevant for defining the projection basis.
Thus we set:
\begin{equation}
\bar{P}_K \coloneq Q_K,
\qquad
\bar{P}_V \coloneq Q_V.
\label{eq:dkv_bases}
\end{equation}
Ranks dimension is then applied by truncation as in Eq.~\ref{eq:dkv_trunc_def}.
\paragraph{Head sharing.}\label{par:head_sharing}
\methodname stores the cache per KV head.
We keep a single shared key basis per layer (i.e., one $\bar P_K$) and share it across heads, while we learn value bases per head: each head $h$ has $\bar P_{V,h}$ and its truncation $P_{V,h}^{(r_V)}$.
This choice provides a trade-off between overhead and flexibility, and reflects that keys are reused across multiple query heads in a group, while values interact with head-specific output-projection pathways.
\paragraph{Overheads.}
Importantly, the MLPs $g_\theta$ of Fig. 1 are only employed offline to learn projection bases, and the latter are the only extra parameters added to the model after calibration. Moreover, as in EigenAttention, projections can be folded into the attention linear weights~\cite{saxena2024eigenattention}. Thus, at inference time, \methodname incurs per-layer FLOP count and runtime data movement overheads identical to those of EigenAttention at iso-rank.
\subsection{Training protocol}
Alg.~\ref{alg:stiefattention} summarizes the \methodname training procedure for the basis predictors $g_{\theta_K}, g_{\theta_V}$ for keys and values.
We train each layer independently by calibrating layer $\ell$ on its uncompressed input activations $\mathcal{X}^{(\ell)}$.
Specifically, for each candidate rank, we train keys and values projection predictors $g_{\theta_K}, g_{\theta_V}$ independently by minimizing the relative decoder-layer output error of Eq.~\ref{eq:dkv_obj}.
We train bases following the head-sharing scheme above.

\begin{algorithm}[t]
\caption{\methodname basis training}
\label{alg:stiefattention}
\begin{algorithmic}[1]
{\renewcommand{\algorithmicrequire}{\textbf{Input:}}%
\REQUIRE Decoder layers $\mathcal{L}$; calibration set $\mathcal{X}$; candidate compressed ranks $\mathcal{R}_K,\mathcal{R}_V$.}
{\renewcommand{\algorithmicensure}{\textbf{Output:}}%
\ENSURE Key bases $\{\bar P_K^{(\ell,r_K)}\}_{\ell,r_K}$ and value bases $\{\bar P_{V,h}^{(\ell,r_V)}\}_{\ell,r_V,h}$.}

\FOR{$\ell \in \mathcal{L}$}
    \STATE Record the uncompressed layer-input activations $\mathcal{X}^{(\ell)}$.
    \FOR{$r_K \in \mathcal{R}_K$}
        \STATE Train $g_{\theta_K}^{(\ell,r_K)}$ on $\mathcal{X}^{(\ell)}$ to solve Eq.~\ref{eq:dkv_obj}.
        \STATE Obtain $\bar P_K^{(\ell,r_K)}$ and its truncation $P_K^{(\ell,r_K)}$.
    \ENDFOR
    \FOR{$r_V \in \mathcal{R}_V$}
        \STATE Train $\{g_{\theta_{V,h}}^{(\ell,r_V)}\}_{h=1}^{H_{\mathrm{KV}}}$ jointly on $\mathcal{X}^{(\ell)}$ to solve Eq.~\ref{eq:dkv_obj}.
        \STATE Obtain $\{\bar P_{V,h}^{(\ell,r_V)}\}_{h=1}^{H_{\mathrm{KV}}}$ and their truncations $\{P_{V,h}^{(\ell,r_V)}\}_{h=1}^{H_{\mathrm{KV}}}$.
    \ENDFOR
\ENDFOR
\end{algorithmic}
\end{algorithm}
\subsection{Rank selection}
\label{sec:rank_selection}
\methodname separates basis learning from rank selection.
For each target KV cache ratio $\rho$, ranks are selected sequentially across layers.
At layer $\ell$, the calibration activations $\mathcal{X}^{(\ell)}_\rho$ are obtained by propagating the calibration set through the previously selected compressed layers.
Thus, the measured error surface $\Delta_\ell^\rho(r_K,r_V)$ is local to layer $\ell$, but is evaluated on inputs that already include the effect of earlier compression decisions.

Let $\mathcal{R}_K$ and $\mathcal{R}_V$ denote the candidate key and value ranks, including the full-rank option $d_h$.
Let $B_\ell$ be the remaining KV cache budget before selecting layer $\ell$.
The current per-layer budget is
\begin{equation}
\tau_\ell = \frac{B_\ell}{L-(\ell+1)},
\label{eq:current_layer_budget}
\end{equation}
with initialization $B_1=L\rho$.
Ranks are selected to minimize the layer-output error under this budget, i.e.,
\begin{equation}
\begin{aligned}
(r_K^\ell,r_V^\ell)
&=
\arg\min_{(r_K,r_V)\in\mathcal{R}_K\times\mathcal{R}_V}
\Delta_\ell^\rho(r_K,r_V)\\
& \mathrm{s.t.}\quad 
\frac{r_K+r_V}{2d_h}\le \tau_\ell .
\end{aligned}
\label{eq:sequential_rank_selection}
\end{equation}
After selection, the remaining budget is updated as
\begin{equation}
B_{\ell+1}
=
B_\ell -
\frac{r_K^\ell+r_V^\ell}{2d_h},
\label{eq:budget_update}
\end{equation}
and the compressed layer output is propagated to construct $\mathcal{X}^{(\ell+1)}_\rho$.

This procedure accounts for cross-layer error propagation from earlier layers because each layer is evaluated on activations produced by the previously compressed prefix.
Further details are provided in Appendix~\ref{app:rank_selection}.
\section{Experimental results}

\subsection{Setup}
\label{sec:experiments}
\begin{table*}[t]
\centering
\small
\setlength{\tabcolsep}{5pt}
\renewcommand{\arraystretch}{1.08}
\caption{
End-to-end performance of \methodname on Llama3-8B and Qwen3-8B.
The table reports the requested target ratio $\rho$ and the actual relative KV cache size induced by the selected ranks.
}
\label{tab:main_results}
\begin{tabular}{l c c cc ccc}
\toprule
\multirow{2}{*}{Model}
& \multirow{2}{*}{$\rho$}
& \multirow{2}{*}{KV cache (rel.) $\downarrow$}
& \multicolumn{2}{c}{Perplexity $\downarrow$}
& \multicolumn{3}{c}{Zero-shot Acc. $\uparrow$} \\
\cmidrule(lr){4-5}\cmidrule(lr){6-8}
& & & WikiText & C4 & HellaSwag & PIQA & MMLU \\
\midrule
\multicolumn{8}{l}{\textbf{Llama3-8B}} \\
FP16 & -- & 1.00 & 6.14 & 9.60 & 0.60 & 0.79 & 0.62 \\
\cmidrule(lr){1-8}
\methodname & 0.90 & 0.86 & 7.71 & 11.61 & 0.58 & 0.79 & 0.60 \\
            & 0.80 & 0.78 & 7.82 & 11.99 & 0.57 & 0.78 & 0.59 \\
            & 0.70 & 0.70 & 8.10 & 12.77 & 0.56 & 0.78 & 0.55 \\
            & 0.60 & 0.60 & 9.74 & 15.92 & 0.52 & 0.76 & 0.50 \\
            & 0.50 & 0.50 & 13.82 & 26.38 & 0.44 & 0.71 & 0.32 \\
\midrule
\multicolumn{8}{l}{\textbf{Qwen3-8B}} \\
FP16 & -- & 1.00 & 9.72 & 13.29 & 0.57 & 0.77 & 0.73 \\
\cmidrule(lr){1-8}
\methodname & 0.90 & 0.89 & 11.56 & 15.98 & 0.55 & 0.77 & 0.67 \\
            & 0.80 & 0.82 & 11.85 & 16.61 & 0.53 & 0.76 & 0.66 \\
            & 0.70 & 0.73 & 12.40 & 17.72 & 0.51 & 0.76 & 0.62 \\
            & 0.60 & 0.64 & 13.39 & 20.60 & 0.48 & 0.75 & 0.56 \\
            & 0.50 & 0.56 & 18.12 & 31.70 & 0.43 & 0.71 & 0.40 \\
\bottomrule
\end{tabular}
\end{table*}

We evaluate \methodname on Llama3-8B~\cite{grattafiori2024llama3herdmodels} and Qwen3-8B~\cite{yang2025qwen3technicalreport}, comparing against the uncompressed FP16 model and EigenAttention~\cite{saxena2024eigenattention} under matched KV cache budgets.
For language modeling, we report perplexity on WikiText~\cite{wikitext} and C4~\cite{c4}; for zero-shot evaluation, we report accuracy on HellaSwag~\cite{hellaswag}, PIQA~\cite{piqa}, and MMLU~\cite{mmlu1,mmlu2} using LM Evaluation Harness~\cite{eval-harness}.
Unless stated otherwise, all results use the  sequential budgeted rank selection procedure described in Sec.~\ref{sec:rank_selection}, with bases calibrated on 512 WikiText sequences of length 2048. An ablation against uniform per-layer allocation and the resulting layer-wise rank profiles are reported in Appendix~\ref{app:rank_selection_results}.
For a fair comparison, EigenAttention is rerun under the same calibration data, sequence length, and evaluation protocol.
Additional training and hardware details are provided in Appendix~\ref{app:experimental_setup}.

\subsection{End-to-end evaluation}
\label{sec:results_e2e}

For \methodname, we evaluate multiple target KV cache ratios $\rho$; ranks are selected by the sequential budgeted procedure of Sec.~\ref{sec:rank_selection}.
For EigenAttention, we follow the layer-wise threshold selection procedure of \citet{saxena2024eigenattention}, varying the SVD threshold to obtain different KV cache ratios.

Table~\ref{tab:main_results} reports \methodname results on Llama3-8B and Qwen3-8B as $\rho$ varies.
Because rank choices are discrete, the achieved KV cache ratio can differ slightly from the requested target.
Across both models, reducing the KV cache yields a smooth degradation in perplexity and zero-shot accuracy rather than an abrupt collapse over the evaluated range.

On Llama3-8B, decreasing the achieved KV cache ratio from $0.86$ to $0.50$ increases WikiText perplexity from $7.71$ to $13.82$ and C4 perplexity from $11.61$ to $26.38$.
The degradation is stronger on C4, indicating that broader-domain language modeling is more sensitive to compression.
Zero-shot accuracy follows a similar trend: PIQA remains comparatively robust down to moderate compression, while HellaSwag and especially MMLU degrade more sharply at lower KV cache ratios.

Qwen3-8B shows the same qualitative behavior.
As the achieved KV cache ratio decreases from $0.89$ to $0.56$, WikiText perplexity increases from $11.56$ to $18.12$, while C4 increases from $15.98$ to $31.70$.
Among zero-shot tasks, PIQA again degrades gradually, whereas MMLU is more sensitive to stronger compression, decreasing from $0.67$ to $0.40$ across the evaluated range.
These results indicate that the sequential budgeted selector produces usable compression tradeoffs on two different model families, while preserving the same task-dependent sensitivity pattern.

\begin{figure}[ht]
    \centering
    \includegraphics[width=\linewidth]{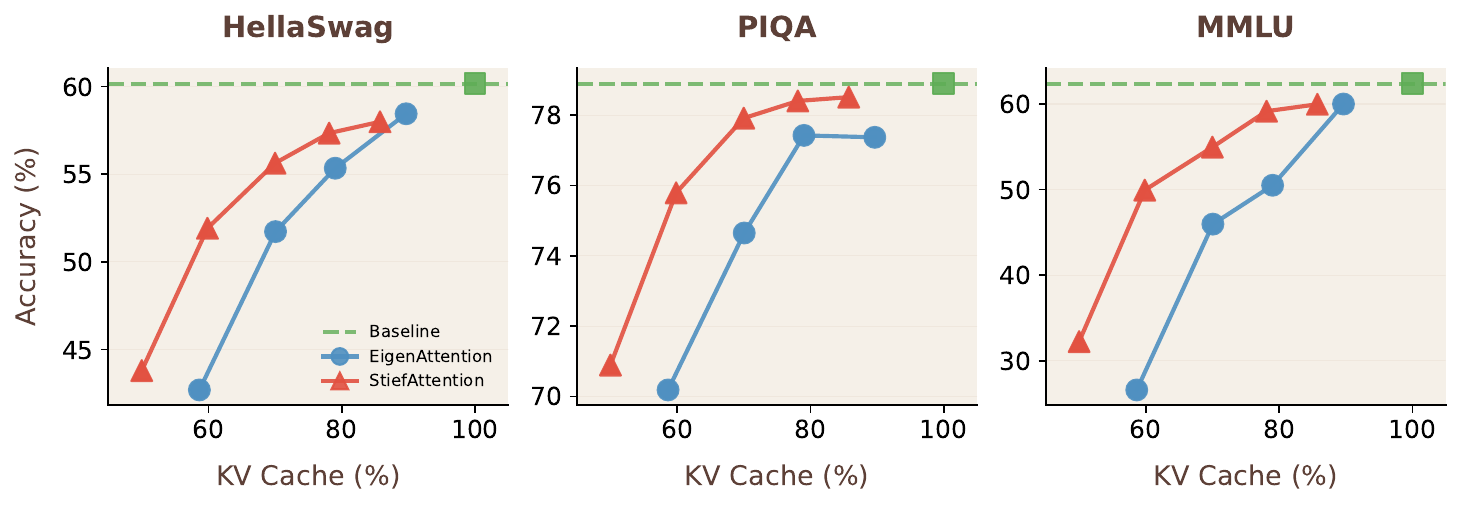}
    \caption{Zero-shot accuracy-memory tradeoff on Llama3-8B.}
    \label{fig:accuracy_llama}
\end{figure}

\begin{figure}[ht]
    \centering
    \includegraphics[width=\linewidth]{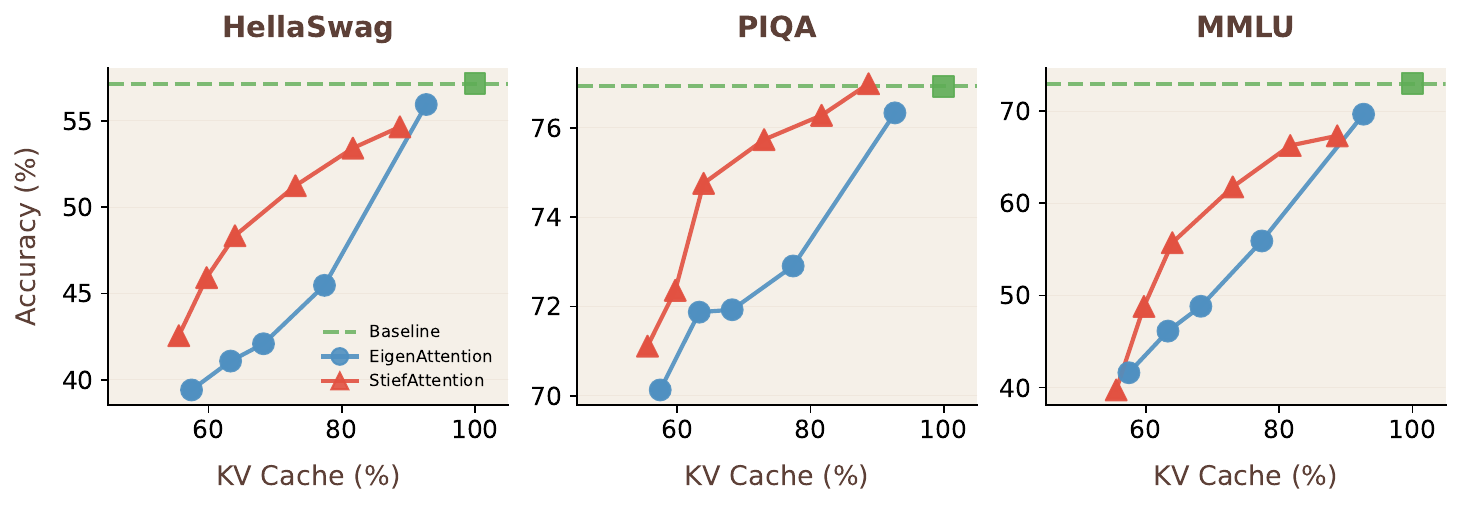}
    \caption{Zero-shot accuracy-memory tradeoff on Qwen3-8B.}
    \label{fig:accuracy_qwen}
\end{figure}

\begin{figure}[ht]
    \centering
    \includegraphics[width=\linewidth]{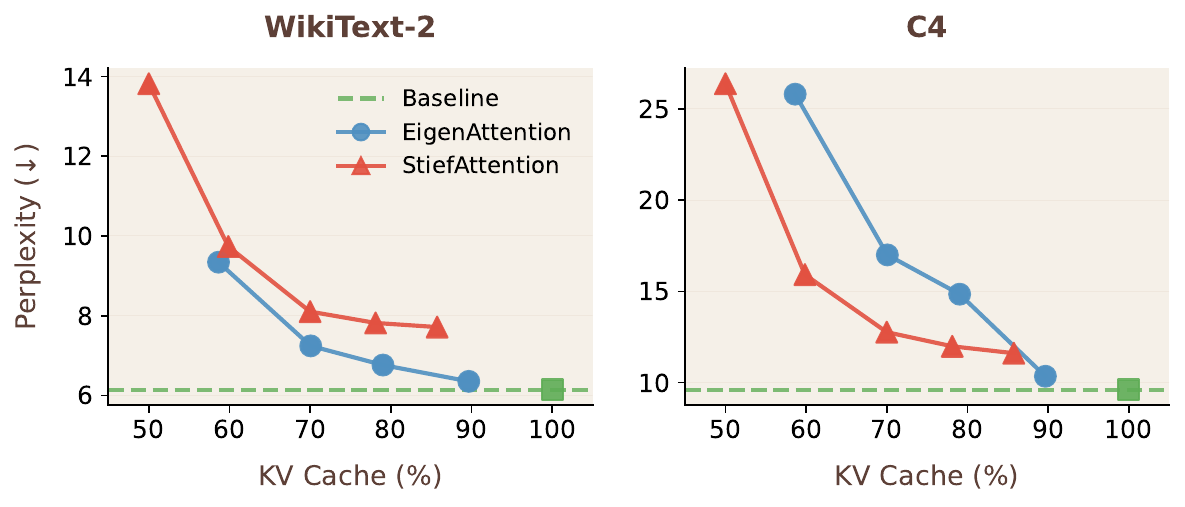}
    \caption{Perplexity-memory tradeoff on Llama3-8B.}
    \label{fig:ppl_llama}
\end{figure}

\begin{figure}[ht]
    \centering
    \includegraphics[width=\linewidth]{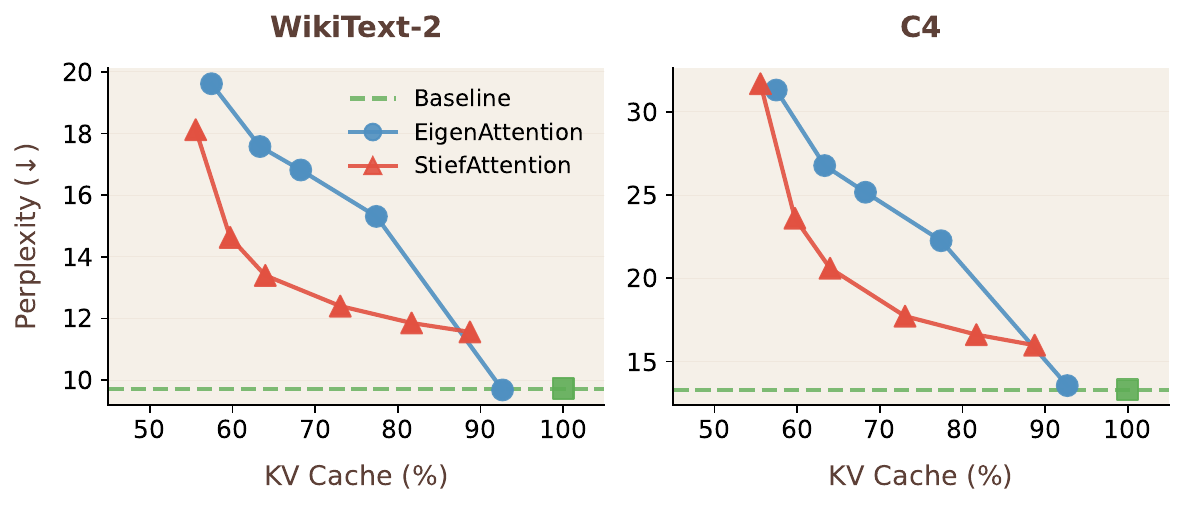}
    \caption{Perplexity-memory tradeoff on Qwen3-8B.}
    \label{fig:ppl_qwen}
\end{figure}

We compare \methodname with EigenAttention in Figs.~\ref{fig:accuracy_llama}--\ref{fig:ppl_qwen} by plotting performance as a function of the achieved KV cache ratio, so methods are compared at comparable memory footprint.

Across both Llama3-8B and Qwen3-8B, \methodname generally achieves a better performance--memory tradeoff than EigenAttention on zero-shot accuracy and language-model perplexity, especially as compression becomes stronger.
On zero-shot benchmarks (Figs.~\ref{fig:accuracy_llama} and~\ref{fig:accuracy_qwen}), \methodname outperforms EigenAttention on HellaSwag, PIQA, and MMLU across most evaluated KV cache ratios, with the largest gains at stronger compression; at very mild compression (KV cache ratio around $0.9$), the two methods can be comparable, and EigenAttention is occasionally slightly better.
On language modeling (Figs.~\ref{fig:ppl_llama} and~\ref{fig:ppl_qwen}), \methodname improves over EigenAttention on C4 across most of the evaluated range, while EigenAttention remains highly competitive on WikiText and can be better at mild compression.
This may reflect stronger alignment between EigenAttention's SVD bases and the WikiText calibration distribution.
Overall, these results indicate that directly optimizing decoder-layer output reconstruction error yields stronger end-to-end robustness at comparable memory footprint, particularly away from the very mild compression regime.

\subsection{Compatibility with KV cache quantization}
\label{sec:results_quantization}

Since projection-based compression reduces the per-head feature dimension, it can be naturally combined with precision reduction.
We apply asymmetric integer quantization to the compressed KV cache tensors of EigenAttention and \methodname, after both methods retain $\approx60\%$ of the original KV cache through low-rank projection.
Following common KV cache quantization practice~\cite{LiuIcml24,hooper2024kvquant}, keys use group-wise scales shared across batch and sequence positions within each KV head, while values use token-wise group scales.

\begin{figure}[t]
    \centering
    \begin{subfigure}[t]{0.48\linewidth}
        \centering
        \includegraphics[width=\linewidth]{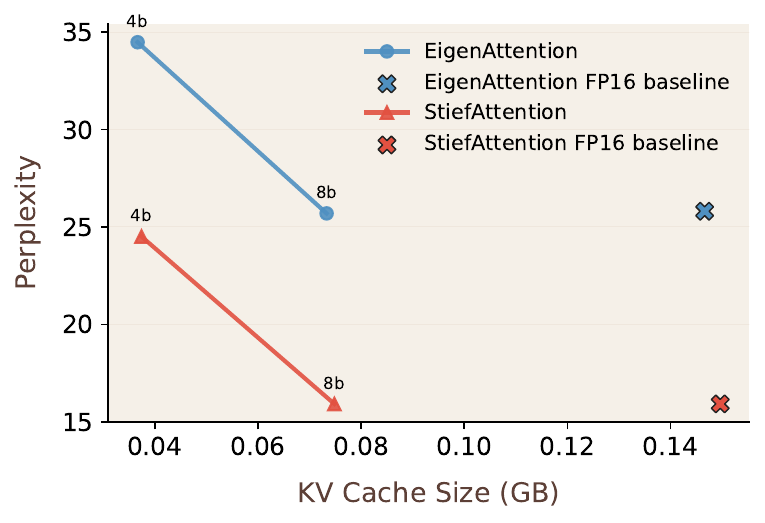}
        \caption{Llama-3-8B}
        \label{fig:quantization_llama}
    \end{subfigure}
    \hfill
    \begin{subfigure}[t]{0.48\linewidth}
        \centering
        \includegraphics[width=\linewidth]{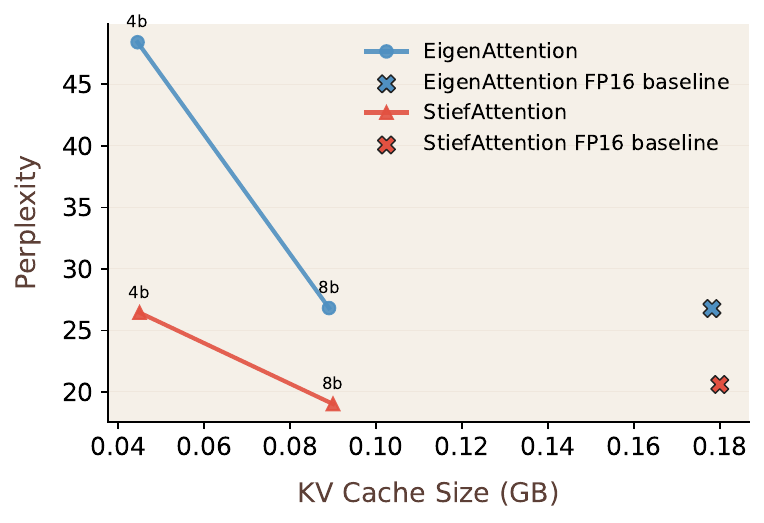}
        \caption{Qwen3-8B}
        \label{fig:quantization_qwen}
    \end{subfigure}
    \caption{KV cache quantization compatibility on C4 at sequence length $2048$ using asymmetric integer quantization with group size $64$.
    EigenAttention and \methodname retain around $60\%$ of the original KV cache before quantization.}
    \label{fig:quantization}
\end{figure}

Figure~\ref{fig:quantization} shows that \methodname is substantially more robust than EigenAttention under low-bit KV cache quantization.
At 8-bit quantization, \methodname remains essentially unchanged relative to its unquantized compressed counterpart, with only negligible perplexity differences.
As the bit-width decreases, \methodname maintains lower perplexity than EigenAttention at comparable KV cache size.
A possible explanation is that optimizing the decoder-layer output objective indirectly penalizes compressed representations that produce large downstream errors, including those caused by poorly scaled or outlier-prone directions.
This may make the resulting KV tensors easier to quantize than bases selected only through a proxy reconstruction criterion.

\subsection{Why \methodname improves end-to-end performance}
\label{sec:results_analysis}
\begin{figure*}[htp!]
    \centering
    \includegraphics[width=\linewidth]{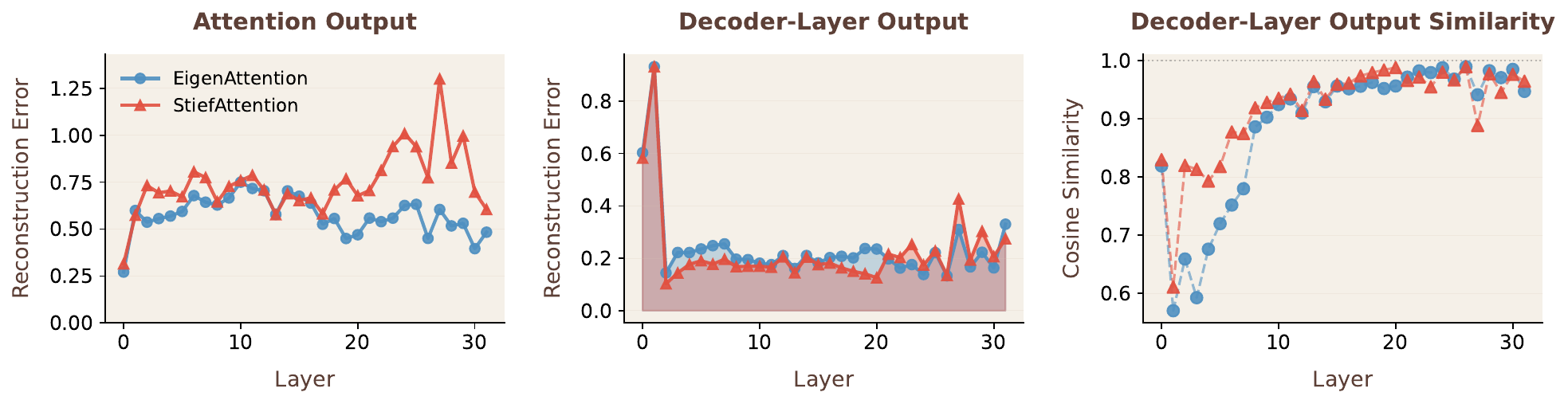}
    \caption{Layer-level output preservation diagnostics. We report (i) attention output reconstruction error, (ii) decoder-layer output reconstruction error $\Delta_\ell$, and (iii) cosine similarity between original and compressed decoder-layer outputs.}
    \label{fig:bases_comparison}
\end{figure*}

We analyze \emph{where} \methodname improves reconstruction by measuring layer-wise output preservation on Llama3-8B on 64 WikiText sequences (2048 tokens). We report results at $(r_K, r_V)=(512,512)$ (half of the original per-head dimension) and observe similar trends across other ranks. For each layer, we log $Q,K,V$, the attention output, and the full decoder-layer output, and compare reconstructions using relative Frobenius-norm error (magnitude) and mean token-wise cosine similarity (direction); Fig.~\ref{fig:bases_comparison} summarizes the results. The corresponding layer-wise diagnostics for Qwen3-8B are reported in Appendix~\ref{app:qwen_basis_analysis}.

As expected, EigenAttention is more robust on proxy targets: it reconstructs intermediate attention quantities more accurately, reducing attention-output error by \textbf{29.6\%} relative to \methodname, with the largest advantage in deeper layers. However, this improvement does not translate to end-to-end behavior. On the decoder-layer output, \methodname achieves lower reconstruction error (\textbf{5.2\%} relative improvement) and, crucially, higher directional agreement (\textbf{+3.3\%} cosine similarity). The cosine gap is most pronounced in early layers: EigenAttention often matches output magnitude but yields activations that are less aligned with the original direction, suggesting that a Frobenius-optimal reconstruction subspace can be misaligned with what downstream nonlinear blocks amplify (softmax/value mixing, residual pathways, normalization, and the MLP).

Overall, these results support the paper’s main claim: proxy reconstruction objectives can accurately preserve intermediate attention quantities while still missing the output-relevant subspace, whereas \methodname directly optimizes decoder-layer outputs and therefore better preserves both magnitude and direction. Importantly, the largest cosine-similarity gains appear in early layers. This matters because perturbations introduced in early layers can be repeatedly transformed by subsequent blocks and thus have a disproportionately large impact on end-to-end behavior~\cite{unreasonable,hc,mhc}. Our diagnostics further indicate that early layers are precisely where SVD-style bases struggle most to preserve output directions.
By training bases against the decoder-layer output metric, \methodname improves alignment in these high-impact layers, which helps explain the consistent end-to-end gains we observe under matched KV cache budgets.
\section{Related Work}
\label{sec:related}

\paragraph{Compressing $K/V$ by reconstruction}
Several methods reduce KV-cache memory by replacing the stored keys/values with low-dimensional projections learned from a calibration set.
Early SVD-style approaches compress keys (and sometimes values) via reconstruction-optimal subspaces, and can be implemented by modifying the KV projection modules so that the cache stores low-rank activations, e.g., ASVD~\citep{asvd}, Palu~\citep{chang2025palu}.
During decoding, the compressed tensors are reconstructed on-the-fly to approximate the original $K/V$, substantially reducing memory occupation with minimal quality loss at modest compression rates.
A key limitation is that reconstruction error is only an indirect proxy for attention and downstream layer behavior, and accuracy can degrade more sharply at higher compression.
ECKVH~\citep{yu2024eckvh} further exploits low-rank structure by grouping KV heads and applying SVD-based compression within each group, but it still relies on reconstruction-style objectives.

\paragraph{Compressing attention interactions}
A key distinction among projection-based methods is \emph{which quantity} the projection is optimized to preserve.
EigenAttention~\citep{saxena2024eigenattention} incorporates queries when constructing the subspace, computing an SVD over concatenated $[K;Q]$ to better preserve query--key geometry under a reconstruction objective.
KQ-SVD~\citep{lesens2025kqsvd} instead targets the pre-softmax interaction matrix, deriving a closed-form low-rank approximation of $QK^\top$ with provable guarantees on score-matrix fidelity.
While these objectives better reflect attention structure than $K/V$ reconstruction alone, they still optimize pre-softmax or intermediate proxies rather than the full decoder-layer output.

\paragraph{Alternative approaches}
Recent works explore designs that better match autoregressive inference constraints or reduce reconstruction overhead.
ZDC~\citep{zhang2024zdc} proposes a zero-delay QKV compression mechanism designed for autoregressive decoding, while TALE~\citep{TALE} introduces token-adaptive low-rank KV-cache approximation and removes explicit reconstruction to reduce overhead.
MatryoshkaKV~\citep{lin2025matryoshkakv} moves beyond fixed SVD bases by fine-tuning orthogonal projections via distillation to better preserve the model's outputs under compression, but it requires a training/distillation stage and still learns projections through a teacher-student objective rather than directly minimizing per-layer output distortion.

In contrast, \methodname learns orthonormal bases that minimize \emph{decoder-layer output} error, rather than fitting bases to reconstruct $K/V$ or approximate $QK^\top$.
\section{Conclusion}
We presented \methodname, a post-training KV cache compression method that learns orthonormal low-rank projection bases by directly minimizing decoder-layer output error, rather than standard proxy objectives based on intermediate attention maps.
Across matched KV cache budgets on Llama3-8B and Qwen3-8B, \methodname improves the performance-memory tradeoff compared to EigenAttention, with particularly strong gains under moderate and strong compression. On Llama3-8B, this includes reducing C4 perplexity by $4.2$ and increasing MMLU accuracy by $8.9$ points at iso-memory.
Our analysis reveals that better preservation of decoder layer outputs, specifically their directions, is more predictive of end-to-end performance than merely reconstructing intermediate attention quantities.

\section*{Limitations}

Our evaluation is still limited in scope.
Due to GPU memory constraints under our two RTX A5000 setup, we mainly report results at sequence length 2048.
Broader evaluation on longer contexts, more datasets, and larger models is needed to assess robustness and scalability.

The current training procedure also simplifies the compression problem.
Key and value bases are calibrated independently for each layer, and rank selection captures cross-layer effects only through a greedy sequential trajectory rather than through exhaustive global optimization.
In addition, the predictor uses only mean and diagonal variance, leaving richer activation summaries such as covariance sketches or low-rank moment features as promising future directions.

We show initial compatibility with KV cache quantization, but other orthogonal compression axes remain unexplored.
Combining \methodname with token pruning, token eviction, or sparse attention could further reduce memory along complementary dimensions.
Future work should also include broader comparisons with recent projection-based KV cache compression systems~\cite{chang2025palu}, ideally under matched KV cache budgets, kernels, and serving measurements.

Finally, a complete system evaluation is still missing.
Future work will analyze prefill and decode latency, throughput, peak memory, and kernel-level overheads under realistic serving conditions.
Lightweight adaptation, such as LoRA fine-tuning after basis selection, may also help recover quality at stronger compression while preserving the post-training nature of the method.

\section*{Societal Impact}

This work aims to improve the memory efficiency of autoregressive LLM inference by reducing KV cache storage and bandwidth requirements. This may lower serving costs and make long-context inference more accessible on memory-constrained hardware. However, efficiency improvements can also reduce the cost of harmful LLM uses, including spam, disinformation, or other misuse. \methodname does not introduce new safeguards and should be combined with the safety, monitoring, and access-control mechanisms required for the underlying models. Since we do not train new foundation models or collect new data, privacy and fairness risks primarily inherit from the pretrained models and benchmark datasets used in evaluation.


\bibliography{bibliography}

\appendix
\section*{Appendix}
\label{sec:appendix}

\section{Additional background and objective details}
\label{app:background_details}

\subsection{Attention, KV cache, and notation}
We consider a transformer decoder layer with attention computed over a prefix of length $n$.
For a per-head dimension $d_h$, let $K,V\in\mathbb{R}^{n\times d_h}$ denote the keys and values for a single attention head over the prefix.
During prefilling, queries are computed for all prefix tokens, so $Q\in\mathbb{R}^{n\times d_h}$.
During autoregressive decoding, the query corresponds to the current token only, so $Q\in\mathbb{R}^{1\times d_h}$.
The per-head attention output is
\begin{equation}
\mathrm{Attn}(Q,K,V)
=
\mathrm{softmax}\!\left(\frac{QK^\top}{\sqrt{d_h}}\right)V.
\label{eq:bg_attn}
\end{equation}

In standard multi-head attention (MHA), each query head has its own key/value head.
Grouped-query attention (GQA) reduces KV memory by letting multiple query heads share the same key/value head, i.e., $H_{\mathrm{KV}}<H_Q$~\citep{gqa}.
The computation in Eq.~\eqref{eq:bg_attn} still applies per query head.

In autoregressive decoding, each layer caches past keys and values to avoid recomputing them for every generated token.
The KV cache size grows linearly with the sequence length, batch size, and number of KV heads, becoming a dominant memory and bandwidth cost at long contexts~\cite{scalingLLM}.

\subsection{SVD notation}
For a matrix $X\in\mathbb{R}^{m\times p}$, we write its SVD as
$X=\mathbf{U}\mathbf{\Sigma}\mathbf{V}^\top$,
where $\mathbf{U}\in\mathbb{R}^{m\times m}$ and
$\mathbf{V}\in\mathbb{R}^{p\times p}$ have orthonormal columns, and
$\mathbf{\Sigma}\in\mathbb{R}^{m\times p}$ is diagonal (rectangular) with nonnegative singular values
$\sigma_1\ge\sigma_2\ge\cdots\ge\sigma_{\min(m,p)}\ge0$.
We denote by $\mathbf{V}_r(X)$ the first $r$ right singular vectors of $X$.

\subsection{Detailed SVD-family objectives}

\paragraph{K-SVD.}
A reconstruction-based baseline chooses an orthonormal basis
$P_K\in\mathbb{R}^{d_h\times r_K}$ to minimize key reconstruction error:
\begin{equation}
\begin{aligned}
&\min_{P_K^\top P_K=I_{r_K}}
\;\;
\|K-KP_KP_K^\top\|_F^2,\\
&P_K=\mathbf{V}_{r_K}(K).
\end{aligned}
\label{eq:app_ksvd}
\end{equation}
The same construction is applied to values by setting
$P_V=\mathbf{V}_{r_V}(V)$.
This objective preserves cached tensors, but does not directly optimize attention behavior or decoder-layer outputs.

\paragraph{EigenAttention.}
EigenAttention constructs
$Z=[K;Q]\in\mathbb{R}^{2n\times d_h}$ and solves
\begin{equation}
\min_{P_K^\top P_K=I_{r_K}}
\;\;
\|Z-ZP_KP_K^\top\|_F^2,
\label{eq:app_eigen}
\end{equation}
using the truncated SVD of $Z$.
This couples keys and queries in the same reconstruction objective, but the resulting subspace is still selected to reconstruct the concatenated tensor, not the decoder-layer output.
For values, EigenAttention applies the reconstruction objective used by K-SVD.

\paragraph{KQ-SVD.}
KQ-SVD targets the pre-softmax interaction matrix by approximating $QK^\top$ directly.
It parameterizes a rank-$r$ approximation as
\begin{equation}
QK^\top
\approx
Q P_Q P_K^\top K^\top
=
(QP_Q)(KP_K)^\top,
\label{eq:app_kqsvd}
\end{equation}
where $P_K,P_Q\in\mathbb{R}^{d_h\times r}$ define the compressed key $K^\downarrow=KP_K$ and compressed query $Q^\downarrow=QP_Q$.
The optimal factors are computed in closed form and are tied to the truncated SVD of $QK^\top$.

For values, KQ-SVD applies the same principle to the value--output pathway by approximating
\begin{equation}
VW_O
\approx
V P_V B^\top W_O,
\label{eq:app_kqsvd_v}
\end{equation}
where $W_O$ denotes the attention output projection.
This selects a value basis that preserves the contribution of values after the output projection, but it remains an intermediate proxy for the full decoder-layer output.

\section{Sequential error-surface construction and budgeted rank selection}
\label{app:rank_selection}

After training the bases in Alg.~\ref{alg:stiefattention}, we construct layer-wise error surfaces for rank selection.
For a target KV cache ratio $\rho$, let $\mathcal{X}^{(1)}_\rho=\mathcal{X}$.
After ranks have been selected for layers $1,\dots,\ell-1$, the inputs to layer $\ell$ are obtained by propagating the calibration set through the previously compressed prefix, yielding $\mathcal{X}^{(\ell)}_\rho$.
The resulting surface $\Delta_\ell^\rho(r_K,r_V)$ is therefore evaluated on inputs that already include the effect of earlier compression decisions.

Let $\mathcal{R}_K$ and $\mathcal{R}_V$ denote the candidate key and value ranks, including the full-rank option $d_h$.
Thus, keys and values can independently remain full rank when this better satisfies the current budget.
For each candidate pair, we define the normalized KV cache cost
\begin{equation}
c(r_K,r_V)=\frac{r_K+r_V}{2d_h}.
\label{eq:rank_cost}
\end{equation}
For the full-rank pair, $\Delta_\ell^\rho(d_h,d_h)=0$ and $c(d_h,d_h)=1$.

\begin{algorithm}[t]
\caption{Sequential error-surface construction and rank selection}
\label{alg:error_surface}
\begin{algorithmic}[1]
{\renewcommand{\algorithmicrequire}{\textbf{Input:}}%
\REQUIRE Layers $\mathcal{L}$; calibration set $\mathcal{X}$; trained bases over $\mathcal{R}_K,\mathcal{R}_V$; target KV cache ratio $\rho$.}
{\renewcommand{\algorithmicensure}{\textbf{Output:}}%
\ENSURE Selected ranks $\{(r_K^\ell,r_V^\ell)\}_{\ell=1}^{L}$ and error surfaces $\{\Delta_\ell^\rho\}_{\ell=1}^{L}$.}

\STATE $\mathcal{X}^{(1)}_\rho \gets \mathcal{X}$
\STATE $B_1 \gets L\rho$
\FOR{$\ell=1,\dots,L$}
    \STATE $\tau_\ell \gets B_\ell/(L-\ell+1)$
    \FOR{$(r_K,r_V)\in\mathcal{R}_K\times\mathcal{R}_V$}
        \STATE Use $P_K^{(\ell,r_K)}$ if $r_K<d_h$; otherwise keep keys full rank.
        \STATE Use $\{P_{V,h}^{(\ell,r_V)}\}_{h=1}^{H_{\mathrm{KV}}}$ if $r_V<d_h$; otherwise keep values full rank.
        \STATE Evaluate $\Delta_\ell^\rho(r_K,r_V)$ on $\mathcal{X}^{(\ell)}_\rho$ using Eq.~\ref{eq:dkv_delta}.
    \ENDFOR
    \STATE Select
    \[
    (r_K^\ell,r_V^\ell)
    =
    \arg\min_{\substack{(r_K,r_V)\in\mathcal{R}_K\times\mathcal{R}_V\\
    c(r_K,r_V)\le \tau_\ell}}
    \Delta_\ell^\rho(r_K,r_V).
    \]
    \STATE $B_{\ell+1}\gets B_\ell-c(r_K^\ell,r_V^\ell)$
    \STATE Propagate layer $\ell$ with ranks $(r_K^\ell,r_V^\ell)$ to obtain $\mathcal{X}^{(\ell+1)}_\rho$.
\ENDFOR
\end{algorithmic}
\end{algorithm}

The selector is greedy: at each layer, it chooses the lowest-error rank pair whose normalized KV cache cost does not exceed the current average remaining budget $\tau_\ell$.
The budget is then updated and the compressed layer output is propagated to construct the activations for the next layer.
Because each $\Delta_\ell^\rho$ is evaluated on activations induced by the previously compressed prefix, the procedure accounts for cross-layer error propagation from earlier layers.
However, earlier rank choices are not revisited after later layers are evaluated; exhaustive end-to-end search over all global rank configurations is computationally prohibitive and left to future work.

\section{Experimental setup details}
\label{app:experimental_setup}

We use publicly available pretrained models and benchmark datasets under their respective licenses and intended research-use conditions.
No new dataset is collected, and no human annotation is performed.
All projection bases and rank profiles are obtained post-training from a held-out calibration set of 512 sequences sampled from WikiText~\cite{wikitext}.
Evaluation uses the standard benchmark splits exposed by the corresponding evaluation tools.
Perplexity on WikiText and C4 is computed with sequence length 2048 using the evaluation code of~\cite{anempiricalQwen3}; zero-shot accuracy is computed with LM Evaluation Harness~\cite{eval-harness}.
Due to GPU memory constraints, we use sequence length 2048 for both calibration and evaluation unless stated otherwise.

For \methodname, we train the basis predictor independently for 5 down-projection ranks uniformly spaced between 50\% and 90\% of the per-head dimension $d_h$.
Training runs for up to 50 epochs with early stopping (patience 5, delta $10^{-6}$).
We use batch size 1 for keys and batch size 4 for values, as value bases were empirically more numerically unstable and required the largest stable batch size supported by our hardware.
We optimize with AdamW (learning rate $5\cdot 10^{-3}$, weight decay $10^{-4}$) and a cosine annealing schedule with $T_{\max}$ set to the total number of training steps.
No downstream fine-tuning is performed.

Unless otherwise stated, reported results are from a single run, due to the computational cost of calibrating multiple ranks and layers.
All experiments are run on two NVIDIA RTX A5000 GPUs with 24 GB each.
The total compute budget is approximately 100 GPU-hours.

\section{Rank-selection ablation and layer-wise rank profiles}
\label{app:rank_selection_results}

We compare the sequential budgeted rank-selection procedure used in the main results against a uniform allocation baseline.
The uniform baseline assigns the same per-layer KV budget across depth, whereas our procedure uses the sequential budgeted selector of Sec.~\ref{sec:rank_selection}, allowing ranks to vary across layers and independently deciding whether keys or values should remain full rank.

\begin{figure}[htp!]
    \centering
    \includegraphics[width=\linewidth]{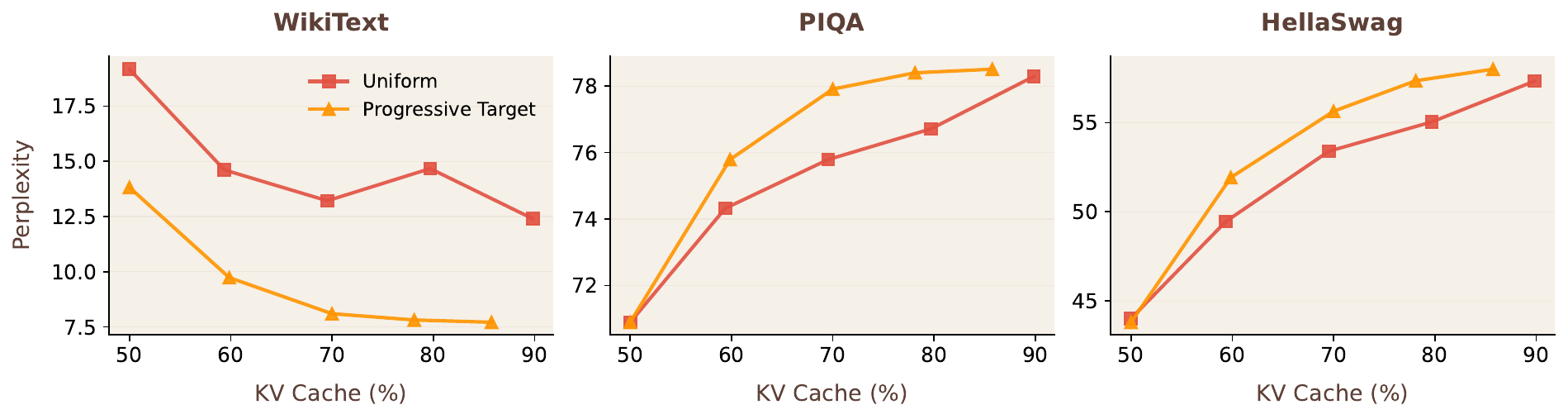}
    \caption{Effect of rank-selection strategy on Llama3-8B.
    We compare uniform per-layer allocation with the sequential budgeted procedure used in the main results, reporting performance as a function of the achieved KV cache ratio.}
    \label{fig:policy_uniform_llama}
\end{figure}

\begin{figure}[htp!]
    \centering
    \includegraphics[width=\linewidth]{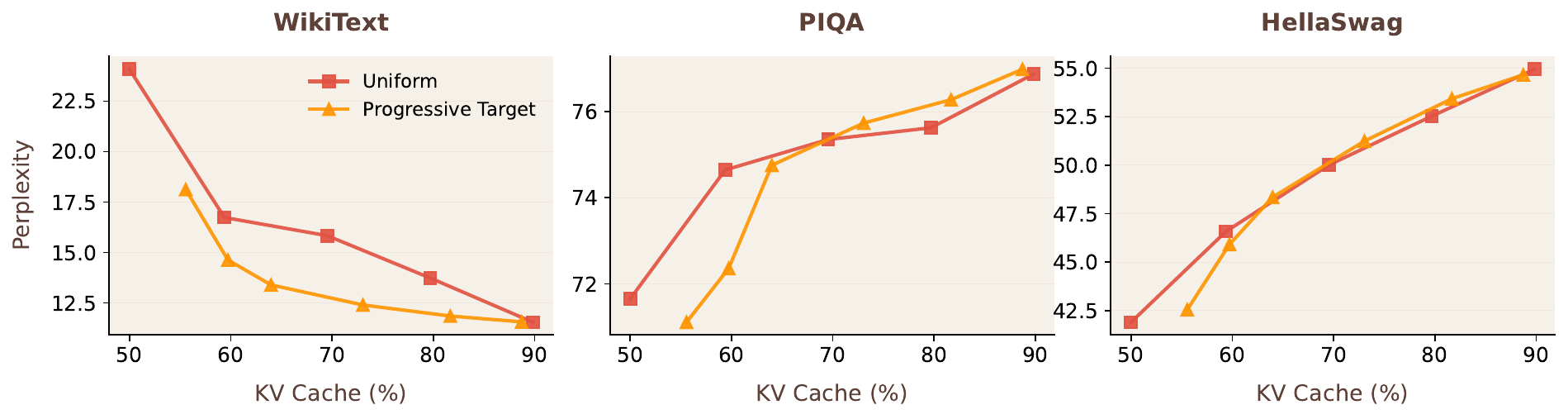}
    \caption{Effect of rank-selection strategy on Qwen3-8B.
    We compare uniform per-layer allocation with the sequential budgeted procedure used in the main results, reporting performance as a function of the achieved KV cache ratio.}
    \label{fig:policy_uniform_qwen}
\end{figure}

Figures~\ref{fig:policy_uniform_llama}--\ref{fig:policy_uniform_qwen} report the end-to-end effect of this choice on Llama3-8B and Qwen3-8B.
Sequential budgeted allocation generally improves the performance--memory tradeoff over uniform allocation, confirming that non-uniform rank allocation is beneficial even when the learned bases are fixed.
Interestingly, at very high compression, the uniform policy can occasionally match or outperform the sequential budgeted selector.
A possible explanation is that, in this regime, the global budget leaves little flexibility: aggressively spending rank on a few sensitive layers may over-compress the remaining layers, while uniform allocation preserves a minimal capacity everywhere.

\begin{figure}[htp!]
    \centering
    \includegraphics[width=\linewidth]{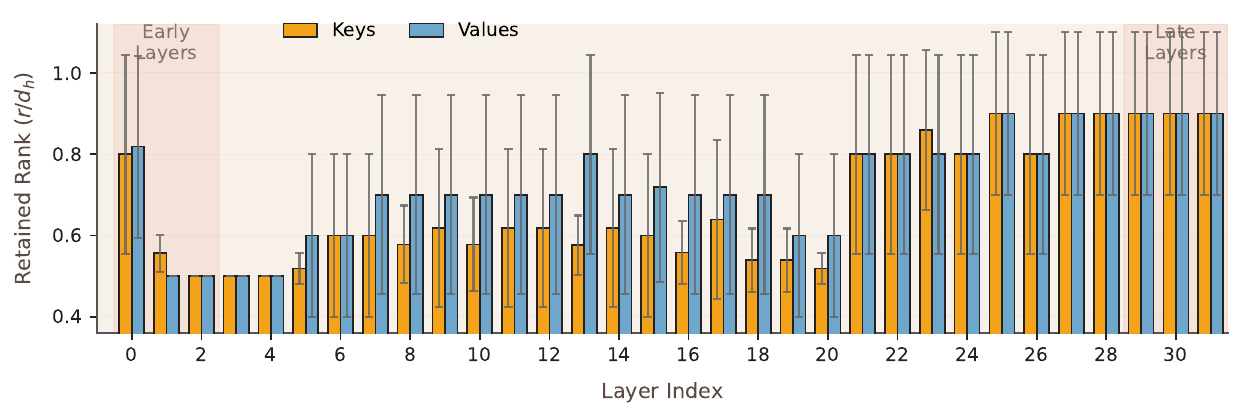}
    \caption{Layer-wise rank profiles produced by sequential budgeted rank selection on Llama3-8B.
    The y-axis reports the normalized retained rank $r/d_h$ for keys and values, averaged across evaluated target KV cache ratios.}
    \label{fig:ranks_llama}
\end{figure}

\begin{figure}[htp!]
    \centering
    \includegraphics[width=\linewidth]{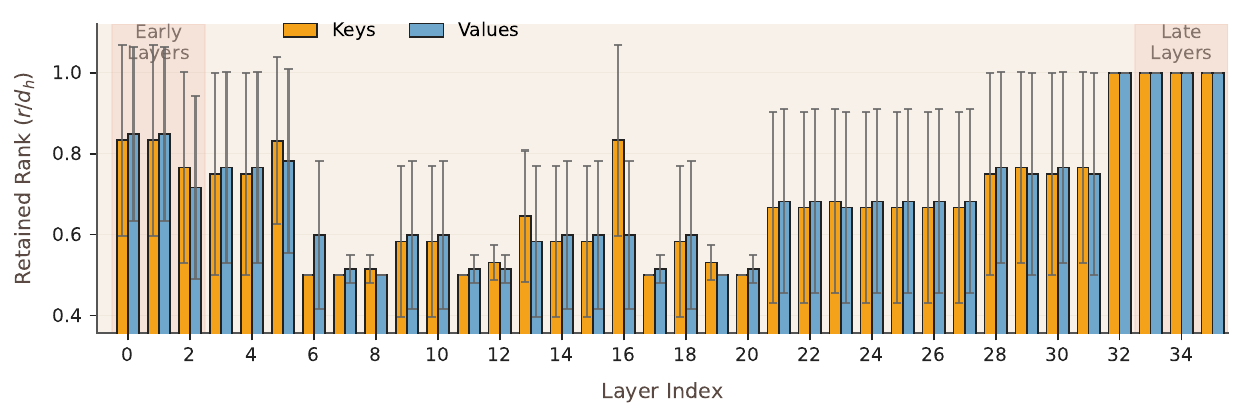}
    \caption{Layer-wise rank profiles produced by sequential budgeted rank selection on Qwen3-8B.
    The y-axis reports the normalized retained rank $r/d_h$ for keys and values, averaged across evaluated target KV cache ratios.}
    \label{fig:ranks_qwen}
\end{figure}

Figures~\ref{fig:ranks_llama}--\ref{fig:ranks_qwen} summarize where the sequential budgeted selector allocates rank.
Across evaluated target KV cache ratios, we record the selected ranks $(r_K,r_V)$ per layer and report their average profile.
The y-axis shows the normalized retained rank $r/d_h$, i.e., the fraction of the original per-head dimension preserved after compression.

Three trends are consistent across Llama3-8B and Qwen3-8B.
First, early layers retain higher ranks, and ranks increase again toward later layers; the latter effect is especially pronounced and may reflect the accumulation of propagated compression error along the sequential calibration trajectory.
Second, values retain a larger fraction of their original dimension than keys, suggesting that the value pathway is harder to compress under the decoder-layer output objective.
Third, intermediate layers show larger variability across budgets, while early and late layers more consistently receive higher ranks.

\begin{figure*}[htp!]
    \centering
    \includegraphics[width=\linewidth]{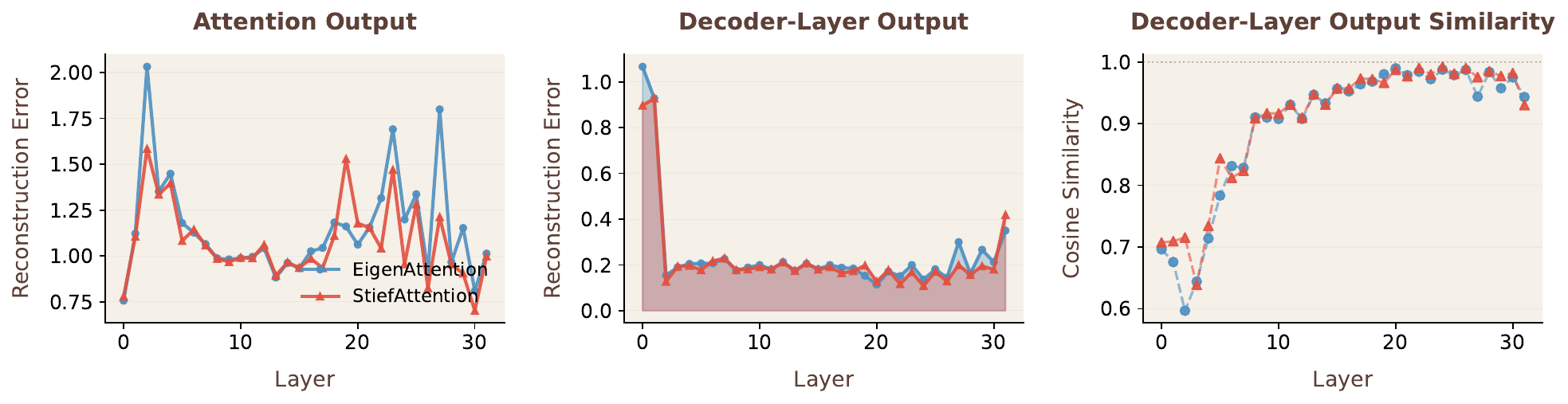}
    \caption{Layer-level output preservation diagnostics for Qwen3-8B. We report (i) attention output reconstruction error, (ii) decoder-layer output reconstruction error $\Delta_\ell$, and (iii) cosine similarity between original and compressed decoder-layer outputs.}
    \label{fig:bases_comparison_qwen}
\end{figure*}

\section{Layer-wise diagnostics on Qwen3-8B}
\label{app:qwen_basis_analysis}

To verify that the layer-level trends observed on Llama3-8B are not model-specific, we repeat the same diagnostic analysis on Qwen3-8B.
As in Sec.~\ref{sec:results_analysis}, we compare EigenAttention and \methodname at fixed rank by measuring attention-output reconstruction error, decoder-layer output reconstruction error, and mean token-wise cosine similarity between original and compressed decoder-layer outputs.

Figure~\ref{fig:bases_comparison_qwen} shows that the qualitative behavior is consistent with the Llama3-8B analysis.
EigenAttention remains competitive on intermediate attention-output reconstruction, but \methodname better preserves the decoder-layer output, especially in terms of directional agreement in early layers.
This supports the claim that optimizing the decoder-layer output objective improves the output-relevant subspace beyond a single model family.
\end{document}